%% file: acl_latex.tex
\documentclass[11pt]{article}

\usepackage[final]{acl}

\usepackage{times}
\usepackage{latexsym}

\usepackage[T1]{fontenc}

\usepackage[utf8]{inputenc}

\usepackage{microtype}

\usepackage{inconsolata}

\usepackage{graphicx}
\usepackage{soul}
\usepackage{amsmath}
\usepackage{booktabs}
\usepackage{multirow}
\usepackage{ulem}
\usepackage{url}
\usepackage[table]{xcolor}
\usepackage{subfigure}
\usepackage[most]{tcolorbox}
\def\model{\textit{text2flow}}

\title{Think Twice Before You Graph: Structured and Semantic Feedback for Robust Graph Generation}
\title{Guiding LLMs with Simulation and Semantic Feedback for Procedural Graph Generation}
\title{Learning to Extract Procedural Graphs with Multiple Cooperative Agents}
\title{Multi-Agent Procedural Graph Extraction with Structural \\and Logical Refinement}

\author{
 \textbf{Wangyang Ying\textsuperscript{1}\thanks{Work done during an internship at NEC Labs America.}},
 \textbf{Yanchi Liu\textsuperscript{2}\thanks{Corresponding author.}},
 \textbf{Xujiang Zhao\textsuperscript{2}},
 \textbf{Wei Cheng\textsuperscript{2}},
\\
 \textbf{Zhengzhang Chen\textsuperscript{2}},
 \textbf{Wenchao Yu\textsuperscript{2}},
 \textbf{Yanjie Fu\textsuperscript{1}},
 \textbf{Haifeng Chen\textsuperscript{2}}
\\
\\
 \textsuperscript{1}Arizona State University,
 \textsuperscript{2}NEC Labs America
\\
 \text{\{wangyang.ying, yanjie.fu\}@asu.edu} \\
 \text{\{yanchi, xuzhao, weicheng, zchen, wyu, haifeng\}@nec-labs.com}
}

\begin{document}
\maketitle

\input{abstract}
\input{introduction}
\input{related_work}
\input{problem_formulation}

\input{method}
\input{experiment}
\input{conclusion}

\bibliography{custom}
\input{appendix_table}
\input{appendix}

\end{document}

%% file: abstract.tex
\begin{abstract}

Automatically extracting workflows as procedural graphs from natural language is promising yet underexplored, demanding both structural validity and logical alignment. While recent large language models (LLMs) show potential for procedural graph extraction, they often produce ill-formed structures or misinterpret logical flows. We present \model{}, a multi-agent framework that formulates procedural graph extraction as a multi-round reasoning process with dedicated structural and logical refinement. The framework iterates through three stages: (1) a graph extraction phase with the graph builder agent, (2) a structural feedback phase in which a simulation agent diagnoses and explains structural defects, and (3) a logical feedback phase in which a semantic agent aligns semantics between flow logic and linguistic cues in the source text.  Important feedback is prioritized and expressed in natural language, which is injected into subsequent prompts, enabling interpretable and controllable refinement. This modular design allows agents to target distinct error types without supervision or parameter updates. Experiments demonstrate that \model{} achieves substantial improvements in both structural correctness and logical consistency over strong baselines. 
\end{abstract}

%% file: introduction.tex
\section{Introduction}
Extracting workflows as structured procedural graphs from natural language documents is a fundamental yet underexplored task~\cite{ProceduralGraph1,ProceduralGraph2}. A procedural graph is defined as a directed graph representing a workflow described by natural language, where nodes correspond to defined procedural elements such as actors, actions, decision points (a.k.a., gateways), constraints, and designated start/end points, and edges denote sequential execution or conditional dependencies among these elements~\cite{herbst1999inductive,maqbool2018comprehensive}. This form of graph explicitly encodes control-flow logic and procedural dependencies, differing from traditional knowledge graph extraction that focuses on identifying entities and relations. Procedural graphs serve as executable structures, enabling downstream applications such as task digitization, automated compliance checking,  and providing structural knowledge input to intelligent systems.

Procedural graphs require modeling not only sequential actions but also complex control structures, such as conditionally executed branches and parallel execution steps. In addition, it requires a comprehensive understanding of the global executability and logical validity of the graphs they produce. However, current methods only partially address these complexities. 
(1) Traditional information extraction techniques, such as entity extraction~\cite{ma-hovy-2016-end,liu2025knowledge} and knowledge graph construction~\cite{ji2021survey,pan2024unifying}, mainly focus on identifying static entities and their semantic relationships, lacking explicit modeling of dynamic execution flows, conditional logic, or iterative loops inherent in procedural documents. 
(2) Rule-based methods or customized neural architectures demonstrated limited generalization beyond predefined, simplified cases~\cite{sholiq2022generating,Bellan}. They fail to adequately capture complex, non-sequential scenarios such as "customers can order only dishes, only drinks, or both," due to their intrinsic limitations in modeling nonlinear and conditional control flows.
(3) Large language models (LLMs) show promise in structured extraction tasks~\cite{zhao2025surveylargelanguagemodels,wang2025efficient}, but current LLM-based methods mainly rely on one-pass zero-shot or few-shot prompting, lacking explicit guidance and iterative refinement. 
Moreover, current self-refinement paradigms~\cite{self-refine-llm,hu2025automated} often rely on heuristic prompting or implicit internal feedback, which offer little guidance when structural flaws block execution paths or when gateway types are subtly inconsistent with textual conditions.
Consequently, these methods struggle with reliably capturing intricate conditional structures and execution logic in procedural graphs~\cite{bellan2022leveragingpretrainedlanguagemodels}.


To overcome the above limitations, in this work, we propose a multi-agent framework, named \textit{\model{}}, for procedural graph extraction via multi-agent structural and logical alignment.
Specifically, we extract two explicit types of external feedback: structural errors derived from multi-path simulation, and logical inconsistencies revealed by comparing execution logic against textual descriptions. 
We decouple the extraction process into three iterative stages: (1) a graph extraction phase with the graph builder agent, (2) a structural feedback phase powered by a simulation agent to detect and explain topological issues, and (3) a logical feedback phase that uses a semantic agent to align between flow logic and the linguistic patterns in the source text. 
Important feedback is prioritized and expressed in natural language, which is injected into the next-round prompt, enabling interpretable and controllable refinement.
By organizing these signals in a unified sandbox and injecting them selectively across iterative reasoning rounds, our framework enables LLMs to revise graphs not only for surface correctness, but with respect to grounded, externalized procedural reasoning. Crucially, although LLM agents are employed to interpret and verbalize feedback, all signals are grounded in observable execution outcomes or explicit inconsistencies between graph logic and text, making them traceable, interpretable, and independent of the generation model’s internal state. This design avoids the compounding hallucinations often observed in self-refinement loops, and enables grounded revisions driven by externally validated failures.
Experiments demonstrate that our approach significantly improves structural correctness and logical consistency over baselines. 

%% file: related_work.tex
\section{Related Work}
Procedural graph extraction research has historically emphasized pipelines that detect events and then link them into sequential relations, often realized through rules or templates, but these approaches rarely generalize beyond linear action flows~\cite{pal2021constructing,lopez2021declarative,ren2023constructing}. Later work introduced non-sequential structures such as branches and optional steps, though coverage and robustness remain limited~\cite{Bellan,honkisz2018concept,epure2015automatic}. Constraint modeling has also been fragmented: data constraints were explored without corresponding attention to action-level dependencies~\cite{friedrich2011process}. Related efforts have additionally framed procedural understanding through representation-centric pipelines, where extracted attributes are transformed into tabular features and optimized via feature selection or learning-based ranking, which improves local predictions but still relies on predefined representations and struggles with global control-flow consistency~\cite{gong2025evolutionary,ying2025data,ying2024revolutionizing,gong2025neuro}. Small-sample neural models exist but raise doubts about scalability. To systematically evaluate these shortcomings, PAGED~\cite{ProceduralGraph2} offers a benchmark of paired documents and procedural graphs, showing both classical and LLM-based methods still struggle to construct coherent non-sequential structures.

Recent work shifts to generation-oriented formulations of procedural graph construction. Cascading LLMs first identify salient events and then generate temporal, hierarchical, or causal edges as code with iterative refinement~\cite{tan2024cascading,deng20222event}; the Set-Aligning Framework treats edges as an order-invariant set with regularizations to boost recall and zero-shot generalization~\cite{tan2024set}. 
Complementary to these text-centric generation strategies, related approaches explore optimization-based pipelines that operate over structured representations or multivariate state spaces, including reinforcement learning for representation selection or topology optimization~\cite{ying2023self,ying2025topology}, as well as transformer-style architectures designed for structured sequence modeling and generation~\cite{kkbugfree,ying2024feature,ying2024unsupervised}. These methods emphasize controllability and efficiency under fixed structural assumptions, and have shown strong performance on locally defined objectives.
However, such pipelines still depend on predefined representations and lack explicit mechanisms to verify global executability or logical consistency across complex control flows. G-PCGRL further casts graph creation as reinforcement learning over an adjacency matrix under explicit constraints, enabling more controllable valid-graph generation than evolutionary search~\cite{rupp2024g}.

%% file: problem_formulation.tex
\begin{figure*}[t]
\centering
\includegraphics[width=\linewidth]{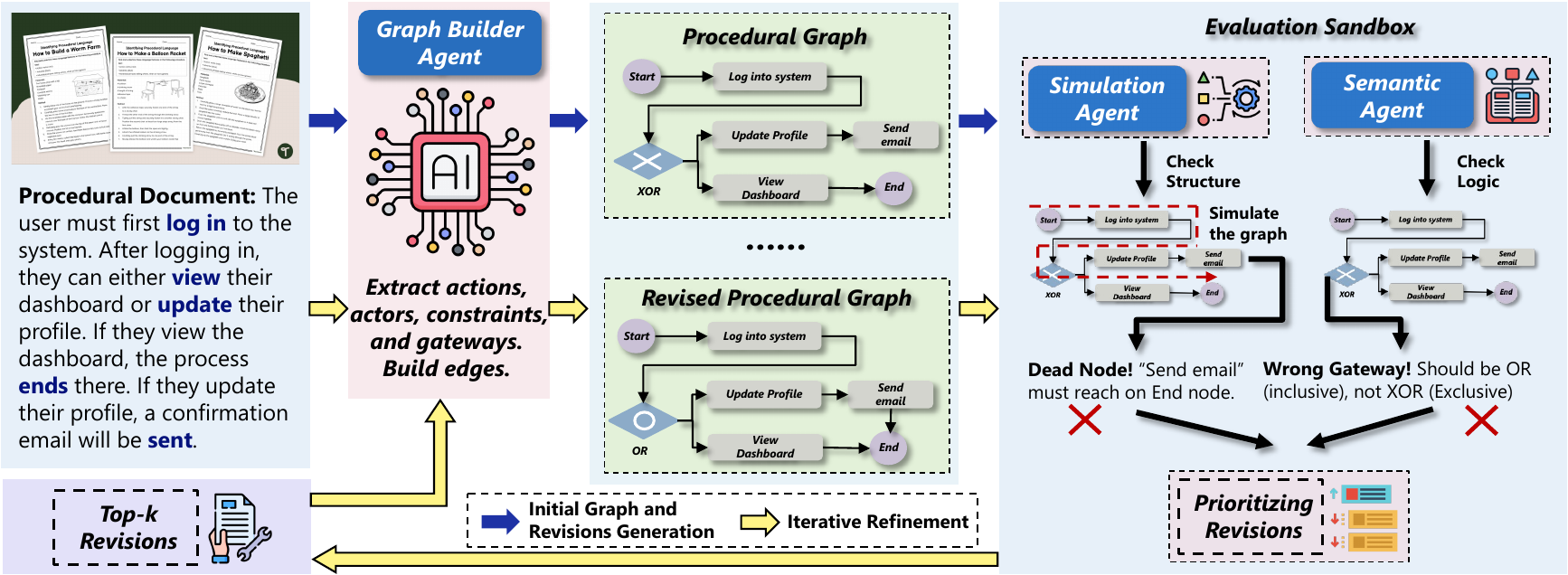}
\caption{Overview of the \model{} multi-agent framework. The Graph Builder Agent generates an initial procedural graph from the document. The Evaluation Sandbox, with simulation and semantic agents, identifies structural and logic errors. Prioritized revisions are iteratively applied to the Graph Builder Agent until a valid graph is produced.}
\label{framework}
\end{figure*}

\section{Problem Formulation}
A procedural graph is a directed process flow graph that represents the control structure of a multi-step procedure, including both sequential and non-sequential execution logic. Given a document, the goal is to generate a graph $G = (V, E)$, where 
\ul{1) $V$ is a set of nodes representing process components}, which includes: \textit{Start} and \textit{End} markers, \textit{Action} nodes that describe concrete procedural steps, \textit{Actor} nodes denoting the responsible entity for each action, \textit{Gateway} nodes that encode control logic (e.g., XOR, OR, AND), and \textit{Constraint} nodes that impose conditions or requirements. 
\ul{2) $E$ is a set of directed edges}, which represents directed flows among these nodes, such as \textit{Sequence Flows} (e.g., action-to-action transitions), \textit{Condition Flows} (e.g., branching conditions at gateways), and \textit{Constraint Flows} (e.g., linking constraints to specific actions).

Our objective is to generate a graph that is both structurally valid and logically faithful to the input document. We formulate this as a multi-round refinement problem, where a multi-agent system iteratively improves the graph under guidance from external structural and logical feedback.

%% file: method.tex
\section{The \model{} Multi-agent Framework}

\subsection{Overview}
\textbf{Figure~\ref{framework}} shows that our framework treats procedural graph extraction as a multi-stage, feedback-driven process, where a large language model (LLM) iteratively generates and refines the output graph under the guidance of structural and logical feedback. Starting from a natural language document, we first use few-shot prompting to elicit an initial procedural graph that captures the core actions, control structures, and execution flows described in the original documents. Recognizing that such initial outputs may contain structural flaws (e.g., disconnected nodes, misused gateways) or logical misalignments (e.g., incorrect interpretation of conditional logic), we introduce a set of lightweight modules that serve as external evaluators. These modules examine the generated graph from two complementary perspectives: structural validity and logical consistency. Each module produces natural language feedback that identifies potential issues and suggests targeted revisions. The LLM then incorporates this feedback in the next round of generation, refining the graph through a loop of diagnosis and revision. By delivering all signals through language, our framework forms a controllable “evaluation sandbox” that enables interpretable, modular, and supervision-free correction of procedural graphs.

\subsection{Graph Builder Agent}
The first step in our framework is to construct an initial procedural graph from the input document using the graph builder agent. Given a natural language paragraph that describes a multi-step procedure, the agent is prompted to output a structured representation of the process as a graph, including the nodes, their types, the control flow edges, and the constraint flow edges.
We adopt a few-shot prompting strategy, where the graph builder agent is provided with a small number of annotated examples in the prompt to guide its generation. Each example consists of a natural language procedure and its corresponding graph represented in a structured format. To make the format LLM-friendly and easy to post-process,  the graph is represented as textual triples, such as `action -> action' or `gateway -> (condition) action'.

The initial graph generated by the agent serves as the first approximation of the underlying procedural structure described in the document. However, due to the complexity of procedural logic and the subtlety of control flow language (e.g., implicit conditions, disjunctions, or parallelism), this initial graph often contains structural errors or misinterpretations of gateway logic. These imperfections motivate the next stage of our framework, where feedback is collected to guide further refinement.

\subsection{Evaluation Sandbox: Structural and Logical Feedback Collection}
To iteratively refine the procedural graph generated by the graph builder agent, we construct an evaluation sandbox that identifies structural and logical issues in natural language. The sandbox operates through two complementary agents: 
(1) a \textbf{simulation agent} that executes the graph to reveal structural failures, 
and (2) a \textbf{semantic agent} that verifies gateway logic based on textual cues. We further define a scoring and selection strategy to prioritize feedback items under prompt constraints.

\subsubsection{Structural Feedback via Simulation Agent}
\noindent\ul{(1) Constructing a simulator to detect the structural issues.} Unlike heuristic-based graph checkers, we treat the generated graph as an executable representation of procedural logic. We design a simulator that mimics multi-path execution from the Start node to the End node, traversing all valid branches under stochastic or enumerative conditions. Each simulation yields a trace-level execution record and flags any violations encountered during traversal. These violations are emitted in a structured form:
\begin{equation}
    \{(\tau, i, c)_k\}_{k=1}^n = \mathcal{S}(G),
\end{equation}
where $\tau$ is the execution trace (sequence of nodes), $i$ is the structured issue encountered (e.g., dead node), $c$ is the condition choice at each gateway in the trace, $n$ is the number of simulations, $\mathcal{S}$ is the notation of the simulator, and $G$ is the graph that needs to be refined. 

\noindent\ul{(2) Using LLM to provide revision feedback.} Each error trace is passed to an LLM-based structural agent, which determines whether the issue reflects a genuine structural flaw and, if so, returns a corresponding repair feedback. This feedback may involve operations such as adding a missing edge, modifying a gateway type, or reconnecting a disjoint node:
\begin{equation}
    f_k = \text{LLM}_{\text{stru}}((\tau, i, c)_k), f_k \in \mathcal{F}_t^{\text{stru}},
\end{equation}
where $t$ is the timestep of the current iteration.

\subsubsection{Logical Feedback via Semantic Agent}
\noindent\ul{(1) Gateway logical feedback.}
Procedural graphs often rely on control flow nodes such as XOR, OR, and AND gateways to represent conditional or parallel behaviors. However, unlike structural errors that manifest during execution, gateway type mismatches arise from a misalignment between graph logic and the linguistic cues in the original text. To support this comparison, we first extract a local textual span for each gateway node, capturing the clause most relevant to its control logic. This step isolates the relevant context and reduces noise from surrounding text. The span is extracted using a semantic retrieval agent:
\begin{equation}
    s_g = \text{LLM}_{extract}(g, G, D).
\end{equation}
where $s_g$ is the textual span corresponding to the gateway $g$, $G$ is the graph that needs to be refined, and $D$ is the original document.

We then ask an LLM to transfer the gateway $g$ and related context back to text description $d_g$. In this way, the comparison of logic between the raw document and the procedural graph is under the same semantic space. Then a semantic consistency agent is used to judge whether the current gateway type matches the logic expressed in $s_g$. Crucially, we only retain feedback when the agent determines that the gateway type is inconsistent with the text, ignoring cases deemed valid or ambiguous: 
\begin{equation}
    f_g = \text{LLM}_{\text{sem}} (s_g, d_g), f_g \in \mathcal{F}_t^{\text{sem}}.
\end{equation}
These feedback items flag specific gateway nodes where the control logic is semantically incorrect.

\noindent\ul{(2) Flow logical feedback.} In addition to the gateways (XOR, OR, AND) which express branching and merging of logic. Flows like sequence, condition, and constraint present logic between nodes.
To capture this flow logic, we segment the entire graph into multiple gateway-to-gateway fragments. For each gateway $g$, we extract its simulation trace $t_g$, defined as the subgraph that spans from $g$ to the closest downstream gateway along the flow sequence. This trace includes: flow-sequence edges along the execution path from gateway $g$ to the next gateway; flow-condition edges that originate from $g$ and point to action nodes within the segment, representing control decisions made by the gateway; flow-constraint edges where constraint nodes point to any action node within this segment, specifying conditional requirements on their execution.

Similarly, we ask an LLM to transfer the trace $t_g$ to a text description $d_t$. In this way, the comparison of logic between the raw document and the procedural graph is under the same semantic space. The semantic consistency agent is used to judge whether the current flow segment matches the logic expressed in the original document. We retain feedback when the agent determines that the gateway segment is inconsistent with the text:
\begin{equation}
    f_t = \text{LLM}_{\text{sem}} (s_t, d_t), f_t \in \mathcal{F}_t^{\text{sem}},
\end{equation}
where $s_t$ is the textual span of trace $t_g$ in the original document.
This feedback highlights segments whose assigned control types are semantically incorrect. 
All of the above feedback is then routed into the sandbox for prioritization and refinement in subsequent iterations.

\subsubsection{Prioritizing High-Impact and Unresolved Feedback}
\noindent\textbf{Why prioritize feedback for refinement?} In each iteration, the LLM-based generator receives a limited number of natural language feedback prompts due to context window and prompt clarity limitations. Presenting too many suggestions simultaneously can overwhelm the model or dilute the guidance signal. Therefore, we prioritize a subset of feedback items that are most likely to yield meaningful corrections. This motivates a scoring-and-selection framework that ranks feedback based on estimated utility.

\noindent\ul{(1) Prioritizing revisions by error frequency impact.} While each simulation provides only a local view of structural issues, the distribution of errors across a large number of simulations reveals their global importance. Structural flaws that appear repeatedly across different execution traces are more likely to lie along central or high-traffic paths in the graph. Fixing such high-frequency errors can unlock access to a broader set of paths and downstream nodes. In contrast, rarely encountered issues may have limited procedural impact even if technically incorrect.

To capture this intuition, we conduct a large number of simulations (e.g., 10,000 trials) using uniform sampling at each gateway to ensure broad path coverage. We then count how frequently each structural issue appears across all wrong simulations. For a given repair suggestion, we define its marginal utility as the normalized frequency of the corresponding error simulation:
\begin{equation}
    u(f_k) = \frac{\text{count}((\tau, i, c)_k)}{\Sigma_{j=1}^n\text{count}((\tau, i, c)_j)}, f_k \in \mathcal{F}_t^{\text{stru}},
\end{equation}
where $\text{count}(\cdot)$ indicates how many times a specific trace was generated in the simulations.

\noindent\ul{(2) Promoting unresolved feedback from previous rounds.} To avoid repeatedly surfacing but ignoring the same issue, we incorporate a RepeatFailure score. A feedback item $f \in \mathcal{F}_t$ is considered a repeat if its semantic meaning is highly similar to a previously surfaced item that was not fixed. We detect repetition using BLEU similarity:
\begin{equation}
    R(f) = \max_{f^{\prime} \in \mathcal{F}_{<t}} [\text{BLEU}(f, f^{\prime})] 
\end{equation}
where $\mathcal{F}_{<t}$ is the feedback in previous iteration.

\noindent\ul{(3) Unified feedback scoring and selection.} While both structural and logical errors affect the overall correctness of the generated graph, structural flaws undermine the executability and interpretability of the graph as a whole. In particular, logic judgments, such as whether a gateway is logically consistent with its textual reference, are only meaningful when the underlying structure is coherent and complete. If a node is unreachable or a branch is disconnected, any attempt to validate its logic becomes unreliable or irrelevant.

Moreover, structural issues can be measured at fine granularity via simulation frequency, yielding a continuous estimate of procedural impact. In contrast, logical errors (e.g., mis-typed gateways) are assessed through discrete judgments and lack trace-level grounding. We encode this asymmetry in the scoring:
\begin{equation}
    w(f) = 
    \begin{cases}
        u(f) + R(f), & f \in \mathcal{F}^{\text{stru}} \\
        R(f), & f \in \mathcal{F}^{\text{sem}}
    \end{cases}
\end{equation}
where $u(f)$ is the simulation-derived impact of structural feedback $f$, $R(f)$ is a shared repair prior, $\mathcal{F}^{\text{stru}}$ denotes structural feedback, and $\mathcal{F}^{\text{sem}}$ denotes logical feedback. 
Consequently, structural issues with broad procedural impact receive higher priority during refinement, while logical feedback is incorporated once the graph is structurally stable enough to support reliable logical comparison.

\noindent\ul{(4) Budget-constrained selection.} We solve a knapsack-style problem to select the top-scoring feedback items within a fixed token budget $B$:
\begin{equation}
    \max_{S \subseteq \mathcal{F}_t} \sum_{f \in S} w(f) \quad \text{s.t.} \quad \sum_{f \in S} \ell(f) \leq B,
\end{equation}
where $\ell(f)$ denotes the token length of feedback $f$. We greedily pick the top-$k$ items by the efficiency score $w(f)/\ell(f)$ under the prompt budget. 

\subsection{Multi-Round Prompting for Graph Refinement}

The feedback-guided graph construction process is conducted in multiple rounds. Each round consists of three stages: (1) generating a procedural graph based on the current prompt, (2) collecting structural and logical feedback via simulation and agent-based analysis, and (3) selecting a prioritized subset of feedback to include in the next prompt. This iterative process enables the model to correct both executional flaws and logical inconsistencies. The refinement process continues for a fixed number of rounds or until no high-impact feedback remains. This multi-round setup allows us to decompose complex graph construction tasks into incremental, feedback-driven revisions, improving structural validity and logical consistency over time.

%% file: experiment.tex
\section{Experiments}
\subsection{Experimental Setup}
\textbf{Dataset Descriptions.}
We conduct our experiments on the PAGED benchmark~\cite{ProceduralGraph2}, a large-scale dataset constructed for procedural graph extraction. The corpus contains 3,394 documents with 37,226 sentences and more than 560K tokens in total. Each document is annotated with diverse process elements. \textbf{Table~\ref{tab:paged-def} in Appendix~\ref{appendix:data}} summarizes the statistics of these core components. The dataset represents procedural knowledge in a text-driven graph format, where sentences are aligned with nodes such as \emph{actions}, \emph{actors}, \emph{gateways}, and \emph{constraints}, and edges are explicitly labeled as flows (e.g., \texttt{`element -> element'} or \texttt{`element -> (condition) element'}) to capture temporal and logical dependencies. In our study, we exclusively utilize the official 20\% held-out test split of PAGED to evaluate model performance. Since our framework operates in a fully unsupervised setting, no training signals from the benchmark are used; instead, we directly apply our method to the test data for evaluation.

\noindent\textbf{Evaluation Metrics.}
The PAGED dataset represents procedural knowledge as tuples in the form of 
\texttt{`element -> element'} or \texttt{`element -> (condition) element'}. Model predictions are evaluated by comparing predicted tuples with gold tuples of the same type.

\noindent\textit{\uline{(1) Tuple Matching.}} For each predicted tuple, we compute its BLEU score~\cite{liang2023knowing} with all gold tuples. The gold tuple with the highest score is selected as the candidate match if the score exceeds 0.5. We then further check:
\begin{itemize}
    \item Start and End Elements: considered matched if its BLEU score exceeds 0.75.
    \item Condition (if present): compared against the gold condition using BLEU.
\end{itemize}

\noindent\textit{\uline{(2) Element-Level Evaluation.}}
\begin{itemize}
    \item Actions, Actors, and Constraints: Evaluated with soft F1~\cite{tandon2020dataset}. A predicted element is correct if it can be matched to a gold element with BLEU above the threshold.
    \item Gateways: Evaluated with hard F1. A predicted gateway is correct only if both (i) its type matches the gold gateway, and (ii) it connects to at least one element that is correctly matched~\cite{dumas2018fundamentals}.
\end{itemize}

\noindent\textit{\uline{(3) Flow-Level Evaluation.}}
\begin{itemize}
    \item A predicted flow is correct if its type and the two connected elements both match the corresponding gold flow.
    \item For Condition Flows, the condition text must also achieve BLEU alignment with the gold condition.
\end{itemize}

\begin{table*}[t]
\centering
\small
\caption{Overall performance. We report the \textbf{F1-score} of each procedural graph component — including action nodes, gateway nodes (XOR/OR/AND), and directed edges — to evaluate the extraction accuracy at both node and structure levels. Higher values indicate better performance. Best results in \textbf{bold}; second-best \ul{underlined}.}
\resizebox{\textwidth}{!}{%
\begin{tabular}{@{}cc|cccccccccc@{}}
\toprule
\multirow{2}{*}{Model}                               &              & \multicolumn{1}{c|}{\multirow{2}{*}{Actor}} & \multicolumn{1}{c|}{\multirow{2}{*}{Action}} & \multicolumn{2}{c|}{Constraint}              & \multicolumn{3}{c|}{Gateway}                               & \multicolumn{3}{c}{Flow}                         \\ \cmidrule(l){5-12} 
                                                     &              & \multicolumn{1}{c|}{}                       & \multicolumn{1}{c|}{}                        & Data           & \multicolumn{1}{c|}{Action} & XOR            & OR             & \multicolumn{1}{c|}{AND} & Sequence       & Condition      & Constraint     \\ \midrule
\multicolumn{2}{c|}{MT-BPMN}                                        & 0.028                                       & 0.308                                        & 0.213          & -                           & 0.485          & -              & 0.279                    & 0.056          & 0.047          & 0.017          \\
\multicolumn{2}{c|}{BRP}                                            & 0.027                                       & 0.276                                        & -              & -                           & 0.469          & -              & 0.337                    & 0.074          & 0.061          & -              \\
\multicolumn{2}{c|}{BPMN-Gen}                                       & -                                           & 0.387                                        & -              & -                           & 0.463          & -              & 0.198                    & 0.091          & 0.022          & -              \\
\multicolumn{2}{c|}{PET}                                            & 0.085                                       & 0.430                                        & 0.069          & -                           & 0.493          & -              & -                        & 0.164          & 0.026          & -              \\
\multicolumn{2}{c|}{CIS}                                            & 0.633                                       & 0.639                                        & -              & -                           & 0.455          & -              & -                        & 0.203          & 0.157          & -              \\ \midrule
\multicolumn{1}{c|}{\multirow{3}{*}{Lamma3.1:8B}}    & Few-Shot  & 0.403                                       & 0.649                                        & 0.534          & 0.151                       & 0.701          & 0.141          & 0.366                    & 0.315          & 0.186          & 0.288          \\
\multicolumn{1}{c|}{}                                & Self-Refine  & 0.568                                       & 0.708                                        & 0.562          & 0.172                       & 0.702          & 0.152          & 0.347                    & 0.368          & 0.204          & 0.399          \\   
\multicolumn{1}{c|}{}                                & Actor-Critic & 0.504                                       & 0.670                                        & 0.556          & 0.165                       & 0.673          & 0.150          & 0.327                    & 0.321          & 0.182          & 0.371          \\
\multicolumn{1}{c|}{}                                & \textbf{\model}         & 0.577                                       & 0.742                                        & 0.644          & 0.178                       & 0.703          & 0.144          & 0.370                    & 0.395          & 0.238          & 0.436          \\ \midrule
\multicolumn{1}{c|}{\multirow{3}{*}{Gemma3:27B}}     & Few-Shot  & 0.621                                       & 0.774                                        & 0.771          & 0.445                       & 0.789          & 0.280          & 0.601                    & 0.371          & 0.276          & 0.477          \\
\multicolumn{1}{c|}{}                                & Self-Refine  & 0.612                                       & 0.774                                        & 0.734          & 0.434                       & 0.784          & 0.239          & 0.642                    & 0.475          & 0.280          & 0.638          \\
\multicolumn{1}{c|}{}                                & Actor-Critic & 0.469                                       & 0.742                                        & 0.625          & 0.339                       & 0.728          & 0.217          & 0.574                    & 0.417          & 0.248          & 0.472          \\
\multicolumn{1}{c|}{}                                & \textbf{\model}         & 0.642                                       & \textbf{0.786}                               & 0.791          & 0.452                       & \textbf{0.839} & \ul{0.345}    & 0.648                    & \ul{0.479}    & \ul{0.342}    & 0.682          \\ \midrule
\multicolumn{1}{c|}{\multirow{3}{*}{Qwen3:30B}}      & Few-Shot  & 0.623                                       & 0.762                                        & 0.816    & 0.421                       & 0.762          & 0.201          & 0.557                    & 0.339          & 0.232          & 0.499          \\
\multicolumn{1}{c|}{}                                & Self-Refine  & 0.653                                       & 0.762                                        & \ul{0.846}    & 0.421                       & 0.792          & 0.201          & 0.587                    & 0.339          & 0.252          & 0.499          \\
\multicolumn{1}{c|}{}                                & Actor-Critic & 0.518                                       & 0.750                                        & 0.714          & 0.388                       & 0.734          & 0.237          & 0.584                    & 0.344          & 0.246          & 0.555          \\
\multicolumn{1}{c|}{}                                & \textbf{\model}         & 0.644                                       & 0.776                                        & 0.843          & 0.411                       & 0.792          & 0.238          & 0.581                    & 0.429          & 0.276          & \ul{0.684}    \\ \midrule
\multicolumn{1}{c|}{\multirow{3}{*}{Mistral3.1:24B}} & Few-Shot  & 0.711                                 & 0.778                                        & 0.776          & 0.461                       & 0.765          & 0.192          & 0.613                    & 0.423          & 0.280          & 0.611          \\
\multicolumn{1}{c|}{}                                & Self-Refine  & \ul{0.717}                                 & 0.778                                        & 0.790          & 0.455                       & 0.804          & 0.249          & 0.639                    & 0.450          & 0.284          & 0.619          \\
\multicolumn{1}{c|}{}                                & Actor-Critic & 0.660                                       & 0.754                                        & 0.709          & 0.335                       & 0.815          & 0.256          & 0.633                    & 0.455          & 0.291          & 0.556          \\
\multicolumn{1}{c|}{}                                & \textbf{\model}         & \textbf{0.723}                              & 0.781                                        & 0.812          & 0.460                       & \ul{0.832}    & 0.282          & \ul{0.677}              & 0.476          & 0.317          & 0.653          \\ \midrule
\multicolumn{1}{c|}{\multirow{3}{*}{GPT-4o}}         & Few-Shot  & 0.684                                       & 0.769                                        & 0.841          & 0.481                & 0.792          & 0.296          & 0.648                    & 0.434          & 0.302          & 0.594          \\
\multicolumn{1}{c|}{}                                & Self-Refine  & 0.691                                       & 0.769                                        & 0.838          & \ul{0.483}                 & 0.796          & 0.299          & 0.673                    & 0.476          & 0.306          & 0.673          \\
\multicolumn{1}{c|}{}                                & Actor-Critic & 0.617                                       & 0.706                                        & 0.760          & 0.441                       & 0.726          & 0.225          & 0.651                    & 0.425          & 0.237          & 0.581          \\
\multicolumn{1}{c|}{}                                & \textbf{\model}         & 0.688                                       & \ul{0.782}                                  & \textbf{0.859} & \textbf{0.485}              & 0.831          & \textbf{0.356} & \textbf{0.692}           & \textbf{0.484} & \textbf{0.357} & \textbf{0.710} \\ \bottomrule
\end{tabular}}
\label{exp:overall_performance}
\end{table*}

\noindent\textbf{Baselines.}
\textbf{(1) Machine Translation-like BPMN (MT-BPMN):}~\cite{Sonbol} which is a semantic transfer-based machine translation approach that generates BPMN models from textual descriptions via intermediate Concept Maps.
\textbf{(2) Beyond Rule-based Process extraction (BRP):}~\cite{Neuberger} which is a data-driven pipeline that extracts process models from text by combining NER, entity resolution, and relation extraction, outperforming rule-based methods.
\textbf{(3) BPMN-Gen:}~\cite{sholiq2022generating} which is a rule-based NLP method for generating BPMN diagrams from textual requirements.
\textbf{(4) PET:}~\cite{Bellan} which uses Roberta-large~\cite{liu2019roberta} as the backbone-model and train the model on the PET dataset\footnote{\url{https://huggingface.co/datasets/patriziobellan/PET}}. A total of 3 epochs are trained, the adopted optimizer is AdamW and the learning rate is set to 5e-6.
\textbf{(5) CIS:}~\cite{bellan2022leveragingpretrainedlanguagemodels} which applies GPT-3 with in-context learning to incrementally extract activities, participants, and control-flow relations from process descriptions via conversational question answering.
\textbf{(6) Self-Refine:}~\cite{self-refine} which introduces an iterative refinement framework where a single LLM alternates between generating outputs, providing feedback on them, and then refining its own responses.
\textbf{(7) Actor-Critic:}~\cite{reflexion}  which frames LLM agents in an actor–critic style, where an actor executes actions and a critic provides verbal feedback to guide iterative self-improvement. This paradigm has become a widely used baseline for multi-agent reasoning and self-correction in LLM research.

\noindent\textbf{Models Details.}
We implement {\model} on 5 LLMs to evaluate the effectiveness. The details of the LLMs are described in \textbf{Appendix~\ref{appendix:model}}.
And the implementation details are described in \textbf{Appendix~\ref{appendix:imp}}. 

\subsection{Overall performance compared with different baselines and LLM models}
This experiment tests whether a multi-agent pipeline with an evaluation sandbox improves procedural-graph extraction over non-LLM parsers and strong LLM baselines across five backbones. \textbf{Table~\ref{exp:overall_performance}} shows that our system attains the highest F1 in most settings, with the largest gains in flow constraints and sequence, and clear gains in gateways follow the semantic consistency in the text. The underlying driver is that the evaluation sandbox provides verifiable signals of two kinds: structural signals from simulation, reachability, and end node checks that reveal broken or missing paths, and logical signals from consistency checks between the text and the candidate gateways. By converting simulation and semantic checks into verifiable constraints, the evaluation sandbox turns free-form generation into guided repair, producing graphs that satisfy global structure and semantics. Compared with Self-Refine, it provides checkable pass/fail signals that indicate what to revise and when to stop; compared with Actor-Critic, specialized critics use simulation traces and text–logic consistency to target gateway typing, reachability, flow constraints, etc. Overall, the experiment shows that injecting external signals enables the LLM to generate procedural graphs that are more accurate and more globally consistent across backbones.

\begin{figure}[t]
\centering
\includegraphics[width=\linewidth]{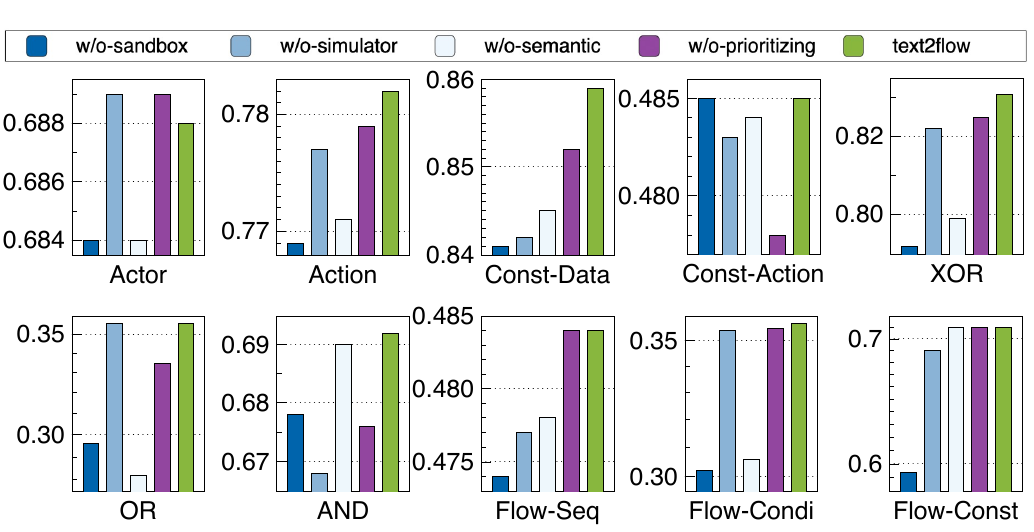}
\caption{Ablation Study on GPT-4o backbone. Each bar shows the performance under different module removals: \textbf{w/o-sandbox} – few-shot generation without evaluation sandbox; \textbf{w/o-simulator} – iterative refinement using only the semantic agent as feedback; \textbf{w/o-semantic} – iterative refinement using only the simulator as feedback; \textbf{w/o-prioritizing} – all revision candidates are sent to the graph builder without filtering. \textbf{\model} denotes the full model with all components enabled.}
\vspace{-0.5cm}
\label{exp:ablation}
\end{figure}

\subsection{Ablation study}
The purpose of this experiment is to examine the contribution of each component in our framework on the GPT-4o backbone by progressively removing the evaluation sandbox, simulator, semantic agent, and prioritization mechanism. As shown in \textbf{Figure~\ref{exp:ablation}}, the full model consistently outperforms all ablated variants across different structural dimensions. Without the sandbox, performance drops notably on control-flow related metrics (e.g., OR, Flow-Condi), highlighting the necessity of external evaluation signals to constrain generation. When relying solely on the simulator (w/o-semantic) or solely on the semantic agent (w/o-simulator), the model suffers on gateway recognition and constraint alignment, suggesting that the two types of feedback are complementary: the simulator enforces structural validity while the semantic agent preserves logical consistency. In addition, disabling the prioritization module (w/o-prioritizing) leads to diminished accuracy on complex node types, since indiscriminate revisions propagate noise and overwhelm the graph builder. These results indicate that each component provides unique benefits, and their integration yields the most stable and robust improvements.

\begin{figure}[t]
\vspace{-0.3cm}
\centering
\includegraphics[width=\linewidth]{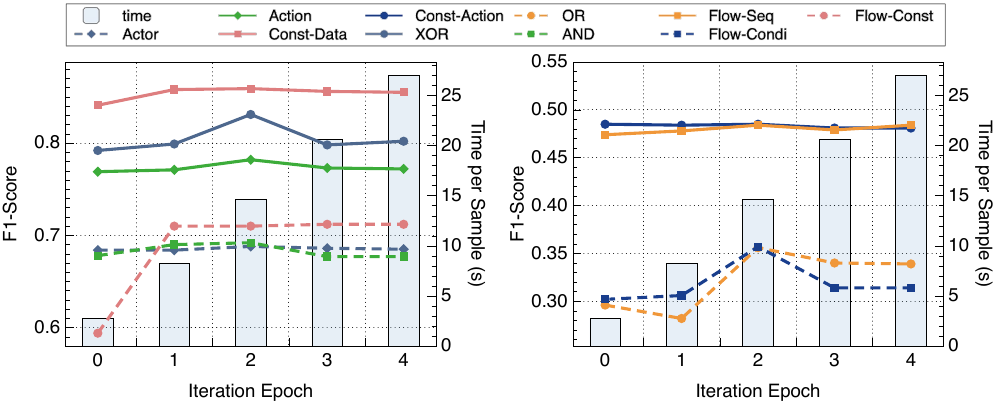}
\caption{F1-scores and average time per sample across iterations (0 = few-shot). Gains peak around 2 iterations, after which improvements plateau while cost increases.}
\vspace{-0.5cm}
\label{exp:iteration}
\end{figure}

\subsection{Performance–Cost Across Iterations}
The purpose of this experiment is to evaluate the scalability of our framework on the GPT-4o backbone by examining how performance and computational cost change with the number of refinement iterations. As shown in \textbf{Figure~\ref{exp:iteration}}, F1-scores across most structural categories improve notably from the few-shot baseline (iteration 0) to the first two refinement rounds, but the gains quickly plateau afterwards. In contrast, the average time per sample grows steadily with more iterations, resulting in diminishing returns. This phenomenon is driven by the fact that early revisions effectively correct major structural and semantic errors, while later rounds tend to revisit already-correct patterns, incurring additional cost without substantial improvement. Consequently, we adopt two iterations as the default setting to strike a balance between accuracy and efficiency.

\begin{figure}[t]
  \centering
  \subfigure[\small GPT-4o]{
    \includegraphics[width=0.225\textwidth]{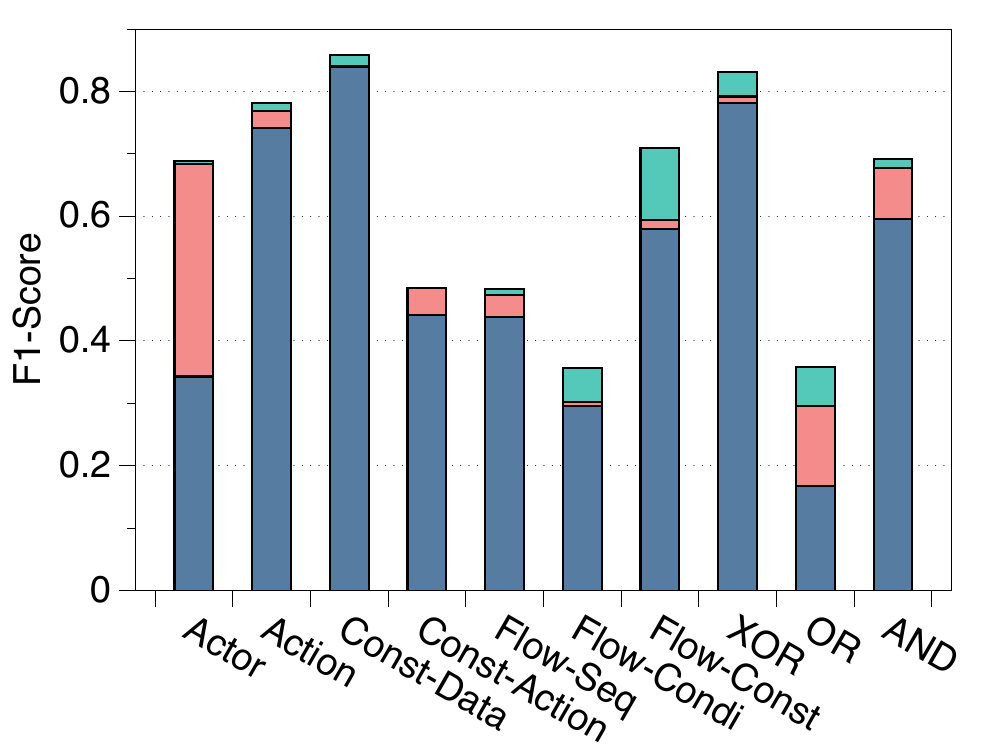}
  }
  \hfill
  \subfigure[\small Qwen3:30B]{
    \includegraphics[width=0.225\textwidth]{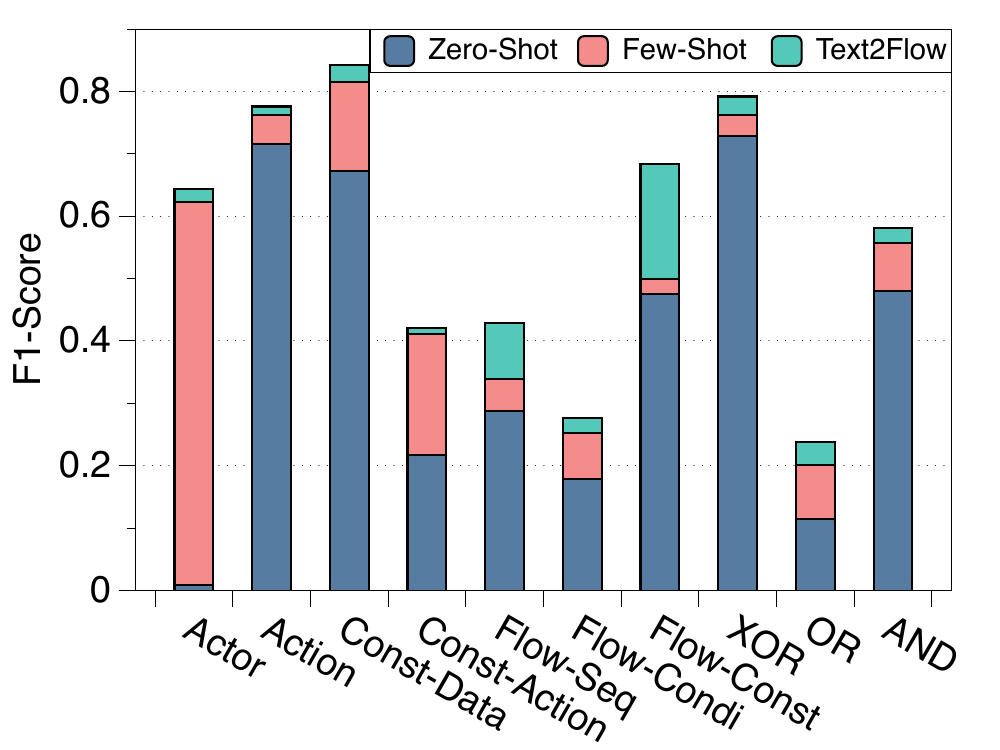}
  }
  \vspace{-0.35cm}
  \caption{F1-score comparison of Zero-Shot, Few-Shot, and {\model} on GPT-4o and Qwen3:30B. {\model} shows consistent gains across all categories.}
  \vspace{-0.5cm}
  \label{fig:improve}
\end{figure}

\subsection{Fine-Grained Comparison with Zero- and Few-Shot Prompts}
To evaluate the effectiveness of {\model}, we conduct a detailed comparison against Zero-Shot and Few-Shot prompting baselines across fine-grained graph components. \textbf{Figure~\ref{fig:improve}} shows the results on GPT-4o and Qwen3:30B.
We observe that {\model} consistently outperforms both Zero-Shot and Few-Shot settings across nearly all components, with especially notable improvements on logically complex or sparsely supervised types such as gateways and flows. These categories often require multi-hop reasoning or global structural consistency, which traditional prompting struggles to capture.
The underlying driver of this improvement lies in our multi-agent feedback loop, which enables iterative correction guided by external structural and semantic constraints. Rather than relying solely on single-pass generation, the system revises outputs based on execution and alignment feedback, leading to more accurate and consistent graph construction.

\begin{table*}[t]
\centering
\small
\caption{Performance comparison of few-shot, reasoning-based, and supervised fine-tuning methods under the Llama-3.1-8B SFT setting for procedural graph extraction.}
\resizebox{\textwidth}{!}{%
\begin{tabular}{@{}cc|cccccccccc@{}}
\toprule
\multirow{2}{*}{Model}                               &              & \multicolumn{1}{c|}{\multirow{2}{*}{Actor}} & \multicolumn{1}{c|}{\multirow{2}{*}{Action}} & \multicolumn{2}{c|}{Constraint}              & \multicolumn{3}{c|}{Gateway}                               & \multicolumn{3}{c}{Flow}                         \\ \cmidrule(l){5-12} 
                                                     &              & \multicolumn{1}{c|}{}                       & \multicolumn{1}{c|}{}                        & Data           & \multicolumn{1}{c|}{Action} & XOR            & OR             & \multicolumn{1}{c|}{AND} & Sequence       & Condition      & Constraint     \\ \midrule
\multicolumn{2}{c|}{Few-Shot}                                        & 0.403                                       & 0.649                                        & 0.534          & 151                           & 0.701          & 0.141              & 0.366                    & 0.315          & 0.186          & 0.288          \\
\multicolumn{2}{c|}{Self-Refine}                                            & 0.568                                       & 0.708                                        & 0.562              & 0.172                           & 0.702          & 0.152              & 0.347                    & 0.368          & 0.204          & 399              \\
\multicolumn{2}{c|}{Actor-Critic}                                       & 0.504                                       & 0.670                                        & 0.556          & 0.165                       & 0.673          & 0.150          & 0.327                    & 0.321          & 0.182          & 0.371          \\
\multicolumn{2}{c|}{\model}                                           & 0.577                                       & 0.742                                        & 0.644          & 0.178                       & 0.703          & 0.144          & 0.370                    & 0.395          & 0.238          & 0.436          \\
\multicolumn{2}{c|}{Fine-tuning}                                            & 0.467                                       & 0.697                                        & 0.527              & 0.187                           & 0.667          & 0.163              & 0.406                        & 0.317          & 0.191          & 0.323              \\ \bottomrule
\end{tabular}}
\label{exp:SFT}
\end{table*}
\subsection{Comparisons with SFT on Procedural Graph Extraction}
Supervised Fine-Tuning with LoRA on Llama-3.1-8B was explored to assess whether task-specific adaptation can improve procedural graph extraction under limited supervision. Experimental results in \textbf{Table~\ref{exp:SFT}} show that fine-tuning yields localized improvements on structurally explicit components, most notably AND/OR gateway typing, where it outperforms text2flow under the same model scale. However, these gains do not translate into consistent improvements across semantic roles or global flow-level metrics, such as Flow-Seq and Flow-Constrain, where fine-tuning remains inferior to inference-based methods. Moreover, the overall performance of the fine-tuned model is still substantially lower than that of larger models (e.g., Gemma-27B or GPT-4o), while introducing nontrivial computational overhead. These results suggest that, given limited annotated data and the semantic complexity of procedural graph extraction, supervised fine-tuning provides partial and complementary benefits, but cannot replace large-model reasoning capabilities.

\begin{table}[]
\centering
\small
\caption{Token Cost on different methods.}
\resizebox{\columnwidth}{!}{%
\begin{tabular}{@{}cccccc@{}}
\toprule
Method     & Zero-Shot & Few-Shot & Self-Refine & Actor-Critic & \model \\ \midrule
Num Tokens & 305       & 1256     & 1801        & 1887         & 2489  \\ \bottomrule
\end{tabular}}
\label{exp:token}
\end{table}
\subsection{Token Cost Analysis}
We further analyze the token cost of different methods by reporting the average prompt and completion token usage per sample. As shown in \textbf{Table~\ref{exp:token}}, token consumption increases substantially with more complex prompting and multi-step reasoning strategies. Zero-shot inference is the most token-efficient (305 tokens), while few-shot prompting already incurs a notable increase (1,256 tokens). Iterative reasoning methods, such as Self-Refine and Actor-Critic, further amplify token usage due to multiple generations and feedback cycles. In contrast, \model exhibits the highest token cost (2,489 tokens per sample), reflecting its reliance on explicit reasoning steps and structured output generation. These results highlight a clear trade-off between performance gains and inference efficiency, suggesting that improvements achieved by advanced reasoning-based methods come at a nontrivial computational and cost overhead.

%% file: conclusion.tex
\section{Conclusion}

In conclusion, this work introduces \model{}, a multi-agent framework for procedural graph extraction that systematically addresses both structural validity and logical consistency, two critical yet often overlooked challenges in this domain. By decoupling extraction, structural refinement, and logical refinement into iterative, agent-driven stages, our approach provides an interpretable and modular solution that enhances reliability without requiring additional supervision or model training. Empirical results demonstrate that this framework yields significant improvements over existing baselines, underscoring the value of multi-round reasoning and feedback-driven refinement. Moreover, we see opportunities to extend this paradigm to broader workflow understanding tasks, integrate domain-specific simulators, and further explore multi-agent collaboration as a general strategy for improving structured reasoning in LLMs.

\section{Limitations}
Our framework is mainly evaluated on structured procedural texts, and its generalizability to informal or domain-specific documents remains to be verified. While the proposed multi-agent refinement process substantially improves structural and logical consistency, it also introduces additional computational overhead compared to single-pass generation. Moreover, the current simulator and semantic agents rely on rule-based heuristics and pretrained LLM judgments, which may not fully capture domain-specific reasoning patterns. Future extensions could incorporate adaptive reward functions or lightweight feedback mechanisms to improve scalability and robustness across diverse procedural domains.

\section{Acknowledgement}
Dr. Yanjie Fu is supported by the National Science Foundation (NSF) via the grant numbers: 2426340, 2416727, 2421865, 2421803.

%% file: appendix_table.tex
\setlength{\tabcolsep}{2mm}
\begin{table*}[!t]
\centering
\caption{Statistics of the dataset. We present statistical information on the number of documents and various elements in the dataset.}
\fontsize{7.8}{9}\selectfont
\begin{tabular}{@{}ccl@{}}
\toprule
Component               & Num    & \multicolumn{1}{c}{Definitions}                                           \\ \midrule
Document                & 3,394   & \multicolumn{1}{c}{-}                                                     \\
Sentence                & 37,226  & \multicolumn{1}{c}{-}                                                     \\
Token                   & 566,639 & \multicolumn{1}{c}{-}                                                     \\ \midrule
Action                  & 36,537  & A concrete operation or step that needs to be executed in the process.    \\
Actor                   & 22,775  & The entity (human, system, or role) responsible for performing an action. \\
Exclusive Gateway (XOR) & 7,024   & A branching point where exactly one of the outgoing paths can be taken.   \\
Inclusive Gateway (OR)  & 1,204   & A branching point where one or more of the outgoing paths may be taken.   \\
Parallel Gateway (AND)  & 2,050   & A branching point where all outgoing paths must be executed in parallel.  \\
Data Constraint         & 3,500   & A specification of data required for an action.                           \\
Action Constraint       & 2,307   & A restriction on how an action should be executed.                        \\
Sequence Flow           & 36,438  & The directed edge that defines the temporal order between elements.       \\
Condition Flow          & 10,598  & A conditional dependency governing process transitions.                   \\
Constraint Flow         & 5,807   & A relation capturing logical or structural restrictions between elements. \\ \bottomrule
\end{tabular}
\label{tab:paged-def}
\end{table*}

%% file: appendix.tex
\appendix
\section{Dataset Descriptions}
\label{appendix:data}
We conduct our experiments on the PAGED benchmark~\cite{ProceduralGraph2}, a large-scale dataset constructed for procedural graph extraction. The corpus contains 3,394 documents with 37,226 sentences and more than 560K tokens in total. Each document is annotated with diverse process elements. \textbf{Table~\ref{tab:paged-def}} summarizes the definitions and statistics of these core components. The dataset represents procedural knowledge in a text-driven graph format, where sentences are aligned with nodes such as \emph{actions}, \emph{actors}, \emph{gateways}, and \emph{constraints}, and edges are explicitly labeled as flows (e.g., \texttt{`element -> element'} or \texttt{`element -> (condition) element'}) to capture temporal and logical dependencies. In our study, we exclusively utilize the official 20\% held-out test split of PAGED to evaluate model performance. Since our framework operates in a fully unsupervised setting, no training signals from the benchmark are used; instead, we directly apply our method to the test data for evaluation.

\section{Model Details}
\label{appendix:model}
\textbf{(1) Llama3.1:8B} is an instruction-tuned model from Meta's LLaMA 3.1 family, designed for efficient reasoning and alignment in downstream tasks.  
\textbf{(2) Gemma3:27B} is a Google-released instruction-following model optimized for structured generation and robust alignment, particularly effective in long-context and knowledge-intensive tasks.  
\textbf{(3) QWen3:30B} is a large-scale model from Alibaba’s Qwen3 family, trained with rich multilingual and domain-specific corpora, achieving strong performance in reasoning and information extraction.  
\textbf{(4) Mistral3.2:24B} is a 24B-parameter model from the Mistral 3.2 series, optimized for efficiency and high-quality instruction following, achieving competitive performance across reasoning and knowledge-intensive tasks. 
\textbf{(5) GPT-4o} is OpenAI’s flagship multimodal model supporting text, vision, and speech, widely regarded as state-of-the-art in general-purpose reasoning and instruction following.

\section{Implementation Details}
\label{appendix:imp}
Our framework supports multiple LLM backbones, including ChatGPT, Gemma, LLaMA, etc.  The initial procedural graph is generated via few-shot prompting with 3 in-context examples. Refinement proceeds for at most 2 iterations (epochs) and terminates early if the evaluation sandbox proposes no further edits. The simulator executes 10,000 trials per graph, sampling each gateway branch with equal probability. At each iteration, we apply no more than three revision suggestions. Open-source models run on 2× NVIDIA RTX A6000 GPUs, while GPT models are accessed via API.

\section{Prompts}

\subsection{Few Shot Graph Generation}
\begin{tcolorbox}[colback=gray!5, colframe=gray!50!black, title=Few Shot Graph Generation Prompt, sharp corners, breakable]
\fontsize{5.5}{7}\selectfont
Please generate procedrual graph based on the extraction rules and the procedural document.\\

\#\#\# DEFINITIONS and RULES:

The Procedural Graph contains the following types of "Nodes" and "Flows":\\

"Nodes":\\
"Start": start node indicates the start of a procedure, represented as "Start". \\
"End": end node indicates the ending of a procedure, represented as "End". \\
"Action": action node indicates a specific step in a procedure, represented as the step itself, such as "prepare the ingredients". \\
"XOR": exclusive gateway, indicates that only one of the following non-sequential actions can be executed, distinguish by numbers, such as "XOR1". \\
"OR": inclusive gateway, indicates that one or more of the following non-sequential actions can be executed, distinguish by numbers, such as "OR1". \\
"AND": parallel gateway, indicates that all of the following actions should be executed in parallel, distinguish by numbers, such as "AND1". \\
"DataObject": DataObject indicates the constraints for the necessary data of the actions, represented as "DataObject(data object)". \\
"TextAnnotation": TextAnnotation indicates essential notices need to be considered for the execution of the actions, represented as "TextAnnotation(essential notices)". \\

" Flows":\\
"SequenceFlow": flow that represents the execution of sequential actions, such as "Start -> prepare the ingredients". \\
"ConditionFlow": the condition flow is used to indicate that the following action is performed under the condition on the Condition Flow, such as "XOR1 -> (condition1) choose the first one".\\
"ConstraintFlow": flow that is used to connect the constraints with corresponding actions, such as "prepare the ingredients -> TextAnnotation(essential notices)".\\

In addition, the actor of corresponding actions is put in the front of corresponding elements to indicate the actor of the following actions if needed, such as "For actor1:".\\

You should generate the graph in the format of "Node -> Node" line by line until generating the whole graph for the given Procedural Document, and keep the text of the nodes and conditions consistent with the original Procedural Document.\\

Here are some examples:\\

\#\# "Procedural Document":\\
Firstly, the customer needs to find an empty seat. If the customer needs dishes, then choose the desired dishes and specify the taste. If the customer needs drinks, then order the drinks and specify the size. The customer then submits the order, which is added to the order list. After enjoying the meal, the customer should choose the payment method. If the credit card is available, the customer pays by credit card; else if the credit card is not available, the customer should pay in cash. For the restaurant, once receiving the order from the order list, it prepares the meal according to the order and prepares the tableware for the customer at the same time. The meal is then served for the customer to enjoy. After that, the restaurant asks the customer to pay for the order and then confirms the payment. Note that the restaurant should provide the receipt if the customer needs. And the procedure ends. \\

\#\# "Procedural Graph":\\

For the customer:\\
Start -> find an empty seat\\
find an empty seat -> OR1\\
OR1 -> (needs dishes) choose the desired dishes\\
OR1 -> (needs drinks) order the drinks\\
choose the desired dishes  -> specify the taste\\
order the drinks -> specify the size\\
specify the taste -> OR2\\
specify the size -> OR2\\
OR2 -> submits the order\\
submits the order -> DataObject(order list)\\
submits the order -> enjoy the meal\\
enjoy the meal -> choose payment method\\
choose payment method -> XOR1\\
XOR1 -> (credit card is available) pay by credit card\\
XOR1 -> (credit card is unavailable) pay in cash\\
pay by credit card -> XOR2\\
pay in cash -> XOR2\\
XOR2 -> End\\
\\
For the restaurant:\\
Start -> receive an order\\
receive an order -> DataObject(order list)\\
receive an order -> AND1\\
AND1 -> prepare the meal\\
AND1 -> prepare the tableware\\
prepare the meal -> AND2\\
prepare the tableware -> AND2\\
AND2 -> serve the meal\\
serve the meal -> ask the customer to pay for the order\\
ask the customer to pay for the order -> confirm the payment\\
confirm the payment -> TextAnnotation(provide the receipt if the customer needs)\\
confirm the payment -> End\\
\\
\#\# "Procedural Document":\\
In the beginning, the staff will receive an order request, and then checks the order type. If the order is standard type, the sufficience of the stock is checked according to the stock table. If the order is special type, upload the order to the factory system. If the stock is sufficient for standard order, the goods will be directly shipped out, else if the stock is insufficient, they will need to be transferred from other warehouses. After that, the staff updates the order status and provide order information to the user. At the same time, the staff needs to bind order information to user account. Finally, the staff record the request status and the procedure ends.\\
\\
\#\# "Procedural Graph":\\
\\
For the staff:\\
Start -> receive an order request\\
receive an order request -> check the order type\\
check the order type -> XOR1\\
XOR1 -> (the order is standard type) check the sufficience of the stock\\
XOR1 -> (the order is special type) upload the order to the factory system\\
check the sufficience of the stock -> DataObject(the stock table)\\
check the sufficience of the stock -> XOR2\\
XOR2 -> (the stock is sufficient) directly shipped out the goods\\
XOR2 -> (the stock is insufficient) transfer the goods from other warehouses\\
directly shipped out the goods -> XOR3\\
transfer the goods from other warehouses -> XOR3\\
XOR3 -> XOR4\\
upload the order to the factory system -> XOR4\\
XOR4 -> AND1\\
AND1 -> update the order status\\
update the order status -> provide order information to the user\\
AND1 -> bind order information to user account\\
provide order information to the user -> AND2\\
bind order information to user account -> AND2\\
AND2 -> record the request status\\
record the request status -> End\\
\\
\#\# "Procedural Document":\\
Start the service by receiving the email from the electronic mailbox, then parse the email content. If the email contains account query request, reply the account information to the user. If the email contains account modification request, record the information needs to be modified. After that, verify the validity of the account and verify the legality of the modified information at the same time if there exists account information to be modified. Otherwise update the verification timestamp of the account directly. Finally, synchronize the email content to the system and the procedure ends.\\
\\
\#\# "Procedural Graph":\\
\\
For the process:\\
Start -> receive the email\\
receive the email -> DataObject(electronic mailbox)\\
receive the email -> parse the email content\\
parse the email content -> OR1\\
OR1 -> (the email contains account query request) reply the account information to the user\\
OR1 -> (the email contains account modification request) record the information needs to be modified\\
reply the account information to the user -> OR2\\
record the information needs to be modified -> OR2\\
OR2 -> XOR1\\
XOR1 -> (there exists account information to be modified) AND1\\
XOR1 -> (otherwise) update the verification timestamp of the account directly\\
AND1 -> verify the validity of the account\\
AND1 -> verify the legality of the modified information\\
verify the validity of the account -> AND2\\
verify the legality of the modified information -> AND2\\
AND2 -> XOR2\\
update the verification timestamp of the account directly -> XOR2\\
XOR2 -> synchronize the email content to the system\\
synchronize the email content to the system -> End \\

\#\#\# Procedural document: \textbf{\textit{\{procedural\_document\}}}
\end{tcolorbox}

\subsection{Structure Check Prompt}
\begin{tcolorbox}[colback=gray!5, colframe=gray!50!black, title=Structure Check Prompt, sharp corners, breakable]
\fontsize{5.5}{7}\selectfont
You are **Structure Checker**, an expert in reviewing procedural graphs.  \\
Your task is to review the Procedural Graph generated by **GraphBuilder** and provide constructive feedback for refinement.\\
\\
\#\#\# DEFINITIONS and RULES:\\
\textbf{\textit{\{extracted\_rules\}}}\\
\\
\#\#\# INPUT:\\
1. The generated Procedural Graph: \textbf{\textit{\{generated\_graph\}}}.\\
2. The original Procedural Document: \textbf{\textit{\{procedural\_document\}}}.\\
3. The Structure Issues from simulator, which lists structural errors or issues detected in the graph: \textbf{\textit{\{structure\_issues\}}}.\\
\\
\#\#\# Structure Feedback Check:\\
- For each *Structure Issue*, carefully check whether it is a real structural problem in the graph.\\
- If the *Structure Issue* is correct, provide a clear, actionable suggestion for how to modify the graph to fix the issue. Prioritize suggestions that add or reconnect nodes/edges, rather than deleting, unless deletion is absolutely necessary.\\
- If you believe a structure-reported issue is not actually a problem, briefly explain why and state that no action is needed for that issue.\\
- If the issue is due to missing information in the document, you may suggest minimal additions to ensure graph connectivity, but avoid inventing unrelated content.\\
- Try to add reasonable nodes, edges, or conditions to the graph to fix the issues, but avoid deleting nodes or edges unless absolutely necessary.\\
- If the structural issue involves auxiliary nodes (such as **DataObject** or **TextAnnotation**), do **not remove** those nodes or their edges. Instead, **retain them for documentation purposes**, and **add a separate edge to reconnect the execution flow to a valid action or decision node**.\\
\\
\#\#\# Important Note on Auxiliary Nodes:\\
- **DataObject** and **TextAnnotation** are auxiliary nodes used for annotation or reference. They are **not part of the executable process flow** and are **ignored in execution traces**.\\
- Edges **pointing to** these nodes (e.g., `Action -> TextAnnotation`) are allowed and useful for documentation. However, they **do not count as valid execution paths**.\\
- If a node only connects to auxiliary nodes, it is still considered a **dead end**. In such cases:\\
    - **Add a new edge to the next action or decision**, **but keep the original auxiliary edge**.\\
    - Do **not** delete the auxiliary node or its connection unless it is incorrect or redundant.\\
- **Do not** use `DataObject` or `TextAnnotation`:\\
    - As starting points in the graph (`DataObject -> Action` is invalid)\\
    - As intermediates in loops, branches, or decisions\\
- The process flow must go through **Action**, **Gateway**, or **Condition** nodes only. Auxiliary nodes may appear in the graph, but only as **side annotations**, not as flow controllers.\\
\\
\#\#\# OUTPUT:\\
If no issues:\\
APPROVED\\
\\
If issues are present, use the following format for each, **don't** reply other information:\\
\\
Issue N\\
- Problem: <copy or summarize the issue>\\
- Status: Confirmed / Not a real issue\\
- Suggestion (if confirmed): <how to fix, ideally by reconnecting, adding or splitting nodes>\\
- Explanation (if not a real issue): <short justification>
\end{tcolorbox}

\subsection{Logic Check Prompt}
\begin{tcolorbox}[colback=gray!5, colframe=gray!50!black, title=Logic Check Prompt, sharp corners, breakable]
\fontsize{5.5}{7}\selectfont
I want you to check whether the gateway segment and its associated text are logically consistent.\\

\#\#\# Input:\\
1. text of gateway segment trace extracted from simulator: \textbf{\textit{\{gateway\_trace\_text\}}}\\
2. original document segment: \textbf{\textit{\{original\_document\}}}\\
\\
\#\#\# Gateway Identification Guidelines\\
\\
1. XOR Gateway (Exclusive Gateway)\\
- Use XOR when only **one** of the possible paths can be taken — the conditions are **mutually exclusive**.\\
- Typical linguistic signals:\\
    - "if..., else if..."\\
    - "if..., however, if..."\\
    - "if..., on the other hand, if..."\\
    - "either A or B"\\
    - "choose one of the following options"\\
    - "if A, skip B"\\
    - "otherwise" / "else"\\
2. OR Gateway (Inclusive Gateway)\\
- Use OR when **one or more** of the paths may be taken independently or together.\\
- Typical linguistic signals:
    - "if..., if..., if..." (without “else” or “otherwise”)\\
    - "also, if..."\\
    - "similarly, if..."\\
    - "you may also..."\\
    - "choose one or more..."\\
    - "any combination of the following"\\
3. AND Gateway\\
- Use AND when **all** following actions must happen, either simultaneously or sequentially.\\
- Typical linguistic signals:\\
    - "at the same time..."\\
    - "meanwhile..."\\
    - "in parallel..."\\
    - "do both A and B"\\
    - "must also do..."\\
    - "simultaneously perform..."\\

Here are some examples:\\
\\
\#\#\# Input:\\
OR1: If there is no information about the old supplier, then check the deadline of 4 business days. Otherwise, continue to do the check.\\
\\
\#\#\# Output:\\
OR1: If there is no information about the old supplier, then check the deadline of 4 business days. Otherwise, continue to do the check.\\
- Status: wrong.\\
- Revision suggestion: Change OR1 to XOR1.\\
- Explanation: The phrase uses a classic XOR pattern — “if..., otherwise...”. These are mutually exclusive conditions: either there is no information about the old supplier, or there is. Only one branch is taken at a time, so this logic requires an XOR, not an OR.\\
\\
Now, I need you to check the following gateways and their corresponding text segments in the Procedural Graph\\
**Only** output the result block if the status is wrong, otherwise, respond with "APPROVED".\\
\\
Only output the result block for the current input using the following format:\\
<Gateway Name>: <text from the document that corresponds to the gateway>\\
- Status: <status of the gateway, correct or wrong>\\
- Revision Suggestion: <suggestion to fix the issue, if any>\\
- Explanation: <explanation of the status, why it is correct or wrong>\\
\end{tcolorbox}

\subsection{Graph Refine Prompt}
\begin{tcolorbox}[colback=gray!5, colframe=gray!50!black, title=Graph Refine Prompt, sharp corners, breakable]
\fontsize{5.5}{7}\selectfont
The Procedural Graph contains the following types of "Nodes" and "Flows":\\
\\
"Nodes":\\
"Start": start node indicates the start of a procedure, represented as "Start".\\
"End": end node indicates the ending of a procedure, represented as "End".\\
"Action": action node indicates a specific step in a procedure, represented as the step itself, such as "prepare the ingredients".\\
"XOR": exclusive gateway, indicates that only one of the following non-sequential actions can be executed, distinguish by numbers, such as "XOR1".\\
"OR": inclusive gateway, indicates that one or more of the following non-sequential actions can be executed, distinguish by numbers, such as "OR1".\\
"AND": parallel gateway, indicates that all of the following actions should be executed in parallel, distinguish by numbers, such as "AND1".\\
"DataObject": DataObject indicates the constraints for the necessary data of the actions, represented as "DataObject(data object)".\\
"TextAnnotation": TextAnnotation indicates essential notices need to be considered for the execution of the actions, represented as "TextAnnotation(essential notices)".\\
\\
"Flows":\\
"SequenceFlow": flow that represents the execution of sequential actions, such as "Start -> prepare the ingredients". \\
"ConditionFlow": the condition flow is used to indicate that the following action is performed under the condition on the Condition Flow, such as "XOR1 -> (condition1) choose the first one".\\
"ConstraintFlow": flow that is used to connect the constraints with corresponding actions, such as "prepare the ingredients -> TextAnnotation(essential notices)".\\
\\
In addition, the actor of corresponding actions is put in the front of corresponding elements to indicate the actor of the following actions if needed, such as "For <actor-name>:". If there is no specific actor mentioned, use "For the process" to indicate the actor of the following actions.\\
\\
You should generate the graph in the format of "Node -> Node" line by line until generating the whole graph for the given Procedural Document, and keep the text of the nodes and conditions consistent with the original Procedural Document.\\
\\
Here are some examples: \textbf{\textit{\{few\_shot\_examples\}}}\\
\\
Previously, another model generated a Procedural Graph, and we have identified several structure issues with it. Please use the list of detected issues and solution suggestions as **references to avoid repeating the same mistakes**.\\
Don't copy the previously generated Procedural Graph, but use it as a reference to generate a new Procedural Graph that is more accurate and complete.\\
\\
\#\#\# "Previously Generated Procedural Graph": \textbf{\textit{\{generated\_graph\}}}\\
\\
\#\#\# "Detected Issues and Solution Suggestions" (you may meet these issues, just refer them as references if available): \textbf{\textit{\{issues\_and\_revisions\}}}\\
\\
Now you need to generate the corresponding Procedural Graph of the following Procedural Document, if there is no specific actor mentioned, use "For the process" to indicate the actor of the following actions:\\
\\
\#\#\# "Procedural Document": \textbf{\textit{\{procedural\_document\}}}
\end{tcolorbox}